\documentclass{article}

\usepackage{arxiv}

\usepackage[utf8]{inputenc} 
\usepackage[T1]{fontenc}    
\usepackage{hyperref}       
\usepackage{url}            
\usepackage{booktabs}       
\usepackage{amsfonts}       
\usepackage{nicefrac}       
\usepackage{microtype}      
\usepackage{lipsum}		
\usepackage{graphicx}
\usepackage{natbib}
\usepackage{doi}
\usepackage{csquotes}

\title{Datasets for Portuguese Legal Semantic Textual Similarity: Comparing weak supervision and an annotation process approaches}

\author{ 
    \href{https://orcid.org/0000-0001-9932-6980}{\includegraphics[scale=0.06]{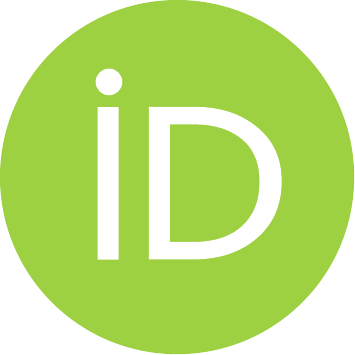}
    \hspace{1mm}Daniel da Silva Junior} \\
	Institute of Computing\\
	Fluminense Federal University\\
	Niterói, RJ 24210-590, Brazil \\
	\texttt{danieljunior@id.uff.br} \\
	\And
	\href{https://orcid.org/0000-0001-9210-1375}{\includegraphics[scale=0.06]{orcid.pdf}\hspace{1mm}Paulo Roberto dos S. Corval} \\
	Law School\\
	Fluminense Federal University\\
	Niterói, RJ 24210-470, Brazil \\
	\texttt{paulocorval@id.uff.br} \\
 \And
	\href{https://orcid.org/0000-0002-9089-7303}{\includegraphics[scale=0.06]{orcid.pdf}
    \hspace{1mm}Aline Paes} \\
	Institute of Computing\\
	Fluminense Federal University\\
	Niterói, RJ 24210-590, Brazil \\
	\texttt{alinepaes@ic.uff.br} \\
  \And
	\href{https://orcid.org/0000-0001-9346-7651}{\includegraphics[scale=0.06]{orcid.pdf}
    \hspace{1mm}Daniel de Oliveira} \\
	Institute of Computing\\
	Fluminense Federal University\\
	Niterói, RJ 24210-590, Brazil \\
	\texttt{danielcmo@ic.uff.br} \\
}

\renewcommand{\shorttitle}{Datasets for Portuguese Legal Semantic Textual Similarity}

\hypersetup{
pdftitle={Datasets for Portuguese Legal Semantic Textual Similarity: Comparing weak supervision and an annotation process approaches},
pdfsubject={cs.LG},
pdfauthor={Daniel da Silva Junior, Paulo Roberto dos S. Corval, Aline Paes, Daniel de Oliveira},
pdfkeywords={Legal Dataset, Semantic Textual Similarity, Data Annotation},
}

\begin{document}
\maketitle

\begin{abstract}
The Brazilian judiciary has a large workload, resulting in a long time to finish legal proceedings. Brazilian National Council of Justice has established in Resolution 469/2022 formal guidance for document and process digitalization opening up the possibility of using automatic techniques to help with everyday tasks in the legal field, particularly in a large number of texts yielded on the routine of law procedures. Notably, Artificial Intelligence (AI) techniques allow for processing and extracting useful information from textual data, potentially speeding up the process. However, datasets from the legal domain required by several AI techniques are scarce and difficult to obtain as they need labels from experts. To address this challenge, this article contributes with four datasets from the legal domain, two with documents and metadata but unlabeled, and another two labeled with a heuristic aiming at its use in textual semantic similarity tasks. Also, to evaluate the effectiveness of the proposed heuristic label process, this article presents a small ground truth dataset generated from domain expert annotations. The analysis of ground truth labels highlights that semantic analysis of domain text can be challenging even for domain experts. Also, the comparison between ground truth and heuristic labels shows that heuristic labels are useful.
\end{abstract}


\section{Introduction}
According to the \textsl{Justice in Numbers} Report 2021 edition\footnote{\label{justicaemnumeros}\url{https://tinyurl.com/bdhbj244}}, the Brazilian Judiciary ended 2020 with 75.3 million cases in progress, of which 25.8 million were new cases opened in the reference year. Among the causes for such many unsolved cases are an insufficient human workforce to meet the demands, and extensive legislation, which has more than 34,000 laws\footnote{\url{https://tinyurl.com/ytzrhc4t}}. Furthermore, Brazil is the sixth most populous country in the world, with an estimated population of 213 million inhabitants in 2020\footnote{\url{https://tinyurl.com/mr33fss7}}, which roughly reflects the number of possible litigants. On the other hand, the \textsl{Justice in Numbers} Report indicates an increase in the productivity of the Brazilian Judiciary, induced by the Judiciary's priority in reducing the ongoing processes number, since if the system continues at this pace, it could take more than 50 years to clear process inventory.

Digitizing the inventory of processes\footnote{\url{https://tinyurl.com/25ep43s8}} is one of the initiatives to relief the judicial system. This digitization also makes possible the use of computational resources that facilitate the analysis of processes and, in some cases, automate repetitive tasks that involve processing a large volume of documents. Automate tasks in the legal context has been supported by Artificial Intelligence (AI) techniques adopted by several legal bodies\footnote{\url{https://tinyurl.com/2v76r4d4}}, including Legal Document Classification\cite{dal2020impact} and Semantic Textual Similarity~\cite{nascimento22}. Primarily, those tasks are addressed with Machine Learning (ML) and Natural Language Processing (NLP) methods.  

Mainly, searching for similar processes is carried out exhaustively in the legal domain, as previous processes can serve as a basis for a new process. The search result is beneficial both for the litigant, who can consider similar processes as a basis for his petition and for the judge to speed up the analysis of the process. It is important to note that this type of search is more effective when considering the textual components of the case, especially when considering the \textit{semantic similarity} between the processes.

Automating tasks in the legal scenario is essential to reduce the stock of unresolved cases, and, therefore, AI can be a great ally in this process. However, experimenting with AI methods and proposing new specific techniques for the legal domain demand the availability of datasets. Furthermore, the automation of particular tasks requires specialized datasets to leverage more sophisticated AI methods. Moreover, many tasks in the legal domain, including similar document retrieval, require \emph{annotated datasets}. However, the annotation task is particularly challenging for the legal domain, requiring subject matter experts who understand the context and vocabulary used to describe the processes.

This article contributes with four Portuguese legal domain datasets, focusing on semantic textual similarity to support similar document retrieval. Two of them, the datasets \emph{TCU Votes}\footnote{TCU = Federal Court of Accounts in Brazil} and \emph{STJ Judgments}\footnote{STJ = Superior Tribunal of Justice in Brazil}, contain the texts and metadata relating to documents extracted from the portals of both bodies but without annotations. The other two, \emph{TCU Votes for Textual Semantic Similarity} and \emph{STJ Judgments for Textual Semantic Similarity}, were generated from the former two datasets, but using a heuristic proposed in this article to annotate documents that are similar to each other. Furthermore, this article presents a ground truth dataset for Semantic Textual Similarity with data from the STJ Semantic Textual Similarity dataset annotated by legal domain experts. This ground truth dataset helped evaluate the heuristic Semantic Textual Similarity dataset and has shown a moderate correlation between the expert and heuristic labels.

The article is organized into five sections in addition to this introduction. Section \ref{sec:related} cites other datasets from the legal domain in Portuguese. Section \ref{sec:corpus} presents the datasets \emph{TCU votes} and \emph{STJ judgments}. Subsequently, Section \ref{sec:corpussst} presents the datasets \emph{TCU Votes for Textual Semantic Similarity} and \emph{STJ Judgments for Textual Semantic Similarity}, as well as the heuristics used for their generation. Section \ref{sec:ground_truth} describes the annotation process for the ground truth dataset, data analysis, and comparison with the heuristic dataset. Finally, Section \ref{sec:conclusion} presents the final considerations.

\section{Related Work} \label{sec:related}

The literature lacks annotated datasets for the Semantic Textual Similarity task with Portuguese legal data. However, some legal datasets contain only {\itshape corpus} of textual data without annotation, and others have annotations to address other specific tasks. The \emph{Iudicium Textum Dataset} \cite{itd_dataset} contains 41,353 documents of the Federal Supreme Court (STF) judgments published between the years 2010 to 2018. De Oliveira \cite{ iceis17} also presents a dataset containing jurisprudence of the Supreme Court of the State of Sergipe, formed by four collections: a) judgments of the Court of Justice (181,994 documents); b) monocratic decisions of the Court of Justice (37,142 documents); c) judgments by Special Courts (37,161 records); and d) monocratic decisions by Special Courts (23,151 documents). Both the \emph{Iudicium Textum Dataset} \cite{itd_dataset} and the {\itshape corpus} provided by De Oliveira \cite{iceis17} are unlabeled datasets.

For the textual classification task, \emph{VICTOR} \cite{luz-de-araujo-etal-2020-victor} is a dataset with more than 692,000 documents from the Federal Supreme Court manually annotated by a team of experts for document type classification tasks and process topic assignment. \emph{LeNER-BR} \cite{de2018lener} is a dataset with 70 documents from judicial courts and Brazilian laws for the named entity recognition (NER) task. Data are annotated with general purpose entities and specific entities of legal knowledge, namely ``Legislation'' for laws and ``Jurisprudence'' for judicial decisions resulting from legal proceedings. \emph{UlyssesNER-Br} was also built for the NER task,  \cite{ulysses2022}, created within the scope of the Chamber of Deputies. It also includes general and specific legal entities, such as ``Fundamental'' and ``Product of Law''. The \emph{UlyssesNER-Br} dataset is divided into two subsets: {\itshape PL-corpus}, with 9,526 publicly available sentence bills, and {\itshape ST-corpus}, private internal documents with 790 sentences of work requests.

\section{TCU Votes and STJ Decisions \itshape{corpora}} \label{sec:corpus}

The first two datasets presented in this article, \emph{TCU votes} and \emph{STJ decisions}, were produced from judgments of the \textit{Superior Tribunal de Justiça (STJ)} and votes from the \textit{Tribunal de Contas da União (TCU)}. The STJ and the TCU are collegiate bodies, that is, bodies where the decision is issued after evaluation and consensus of the responsible members. The \textit{decisions} are texts of judgments of collegiate bodies, which cover only the main points of a discussion. On the other hand, a \textit{vote}, in the context of collegiate bodies, is the exposition, evaluation, and opinion on the decision to be taken for a case in question carried out by the responsible member, called rapporteur\footnote{\url{https: //www.congressonacional.leg.br/legislacao-e-publicacoes/glossario-legislativo/-/legislativo/termo/relator_quanto_ao_papel}}.

The particularity of these data consists of precedents of \textit{jurisprudence} used by the bodies. Jurisprudences are understandings adopted by legal bodies that guide the decision for a given matter. These understandings are formulated by analyzing previous decisions on the same subject -- the precedents -- and aim to standardize decisions and speed up the processes of recurrent matters. 

The texts were obtained from a data scraping routine of the respective organs' websites. After executing the data scraping routine, records with null information or duplicate records were removed. As seen in Table \ref{tab:data_info}, the resulting STJ judgments dataset has a number of records, represented by the line \emph{Decisions}, much higher than the TCU votes dataset, where the total number of records is represented by the line \emph{Votes}, in addition to the superiority of jurisprudence represented. The \ref{tab:data_info} Table also provides information on the categorization used for the data of each body, which are displayed in a hierarchical descending order; that means, in TCU data, a vote has a Subtopic that belongs to a Topic that by turn belongs to an Area.

\begin{table}[ht]
\centering
\caption{Characteristics of the \textit{STJ} and \textit{TCU} data used in the experiments.}
\label{tab:data_info}       
\begin{tabular}{lclc}
\hline
\multicolumn{2}{|c|}{\textbf{TCU}}                        & \multicolumn{2}{c|}{\textbf{STJ}}                          \\ \hline
\multicolumn{1}{|l|}{Votes} & \multicolumn{1}{c|}{371} & \multicolumn{1}{l|}{Decisions}     & \multicolumn{1}{c|}{7403} \\ \hline
\multicolumn{1}{|l|}{Jurisprudences} & \multicolumn{1}{c|}{44} & \multicolumn{1}{l|}{Jurisprudences} & \multicolumn{1}{c|}{1458} \\ \hline
\multicolumn{1}{|l|}{Areas}    & \multicolumn{1}{c|}{4}   & \multicolumn{1}{l|}{Subjects}  & \multicolumn{1}{c|}{7}    \\ \hline
\multicolumn{1}{|l|}{Themes}    & \multicolumn{1}{c|}{27}  & \multicolumn{1}{l|}{Natures} & \multicolumn{1}{c|}{68}   \\ \hline
\multicolumn{1}{|l|}{Sub-theme} & \multicolumn{1}{c|}{38}  & \multicolumn{1}{l|}{}          & \multicolumn{1}{l|}{}     \\ \hline
                               & \multicolumn{1}{l}{}     &                                & \multicolumn{1}{l}{}     
\end{tabular}
\end{table}

The datasets presented in this article are available in \textit{CSV} format at the URL {\url{https://osf.io/k2qpx/}}. The \emph{TCU votes} dataset has the attributes: \texttt{AREA, THEME, SUB-THEME, STATEMENT, PROCESS, YEAR, TYPE\_PROCESS, REPORTER} and \texttt{VOTE}. The \texttt{STATEMENT} attribute defines the jurisprudence to which a VOTE, which is a precedent, is associated. The data set \emph{STJ Judgments} has the attributes \texttt{MATTER, NATURE, THEME, PROCESS, REPORTER, BODY, JUDGMENT\_DATE, PUBLICATION\_DATE} and \texttt{SUMMARY}. In this case, the \texttt{THEME} attribute defines the jurisprudence to which a \texttt{SUMMARY}, which is a precedent, is associated.

\begin{figure}[!htb]
\minipage{0.49\textwidth}

\includegraphics[width=.9\textwidth]{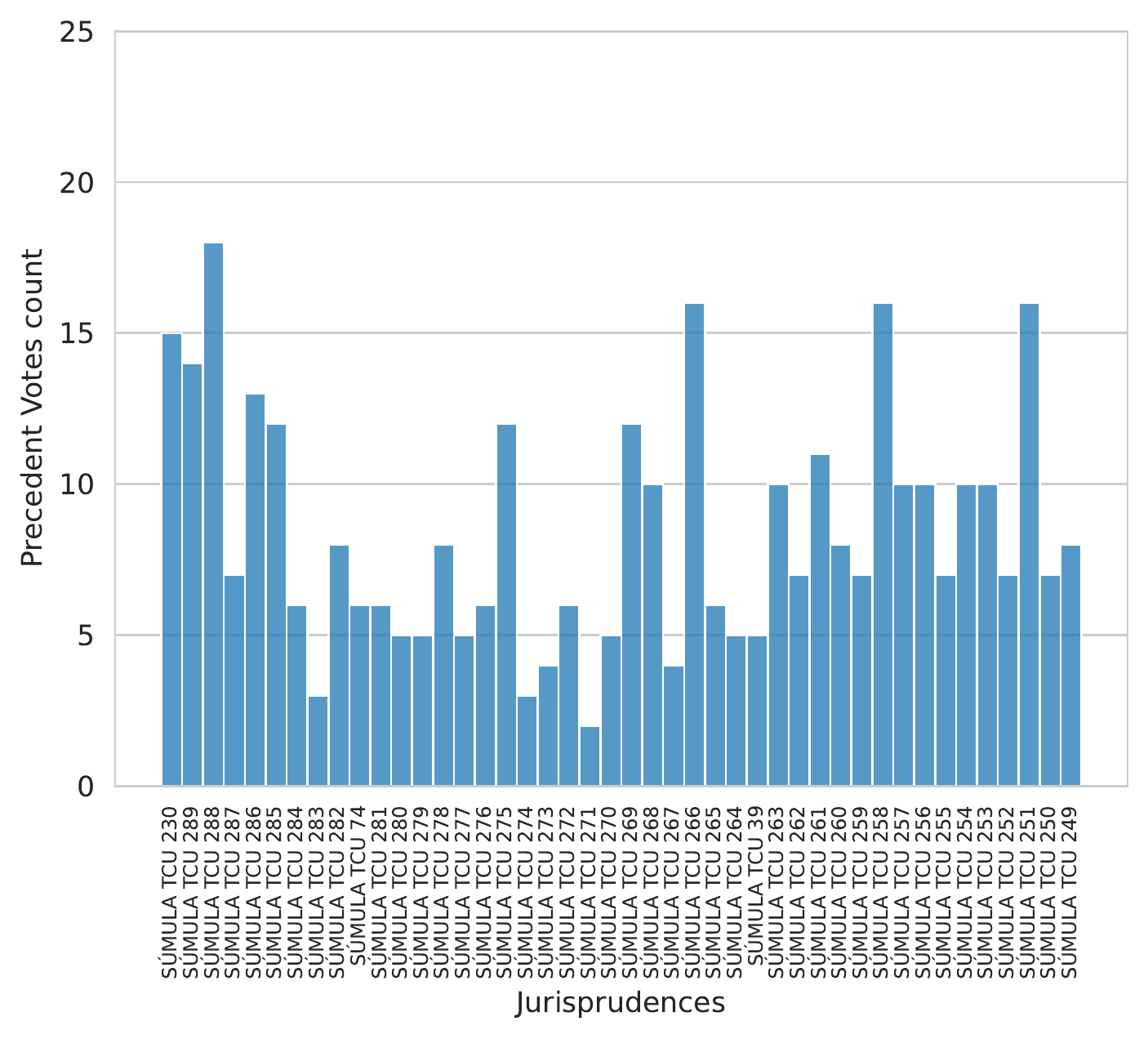}
\caption{Histogram of Precedents x \emph{TCU votes} Jurisprudences}
\label{fig:prec_juris_tcu}

\endminipage\hfill
\minipage{0.49\textwidth}

\includegraphics[width=.9\textwidth]{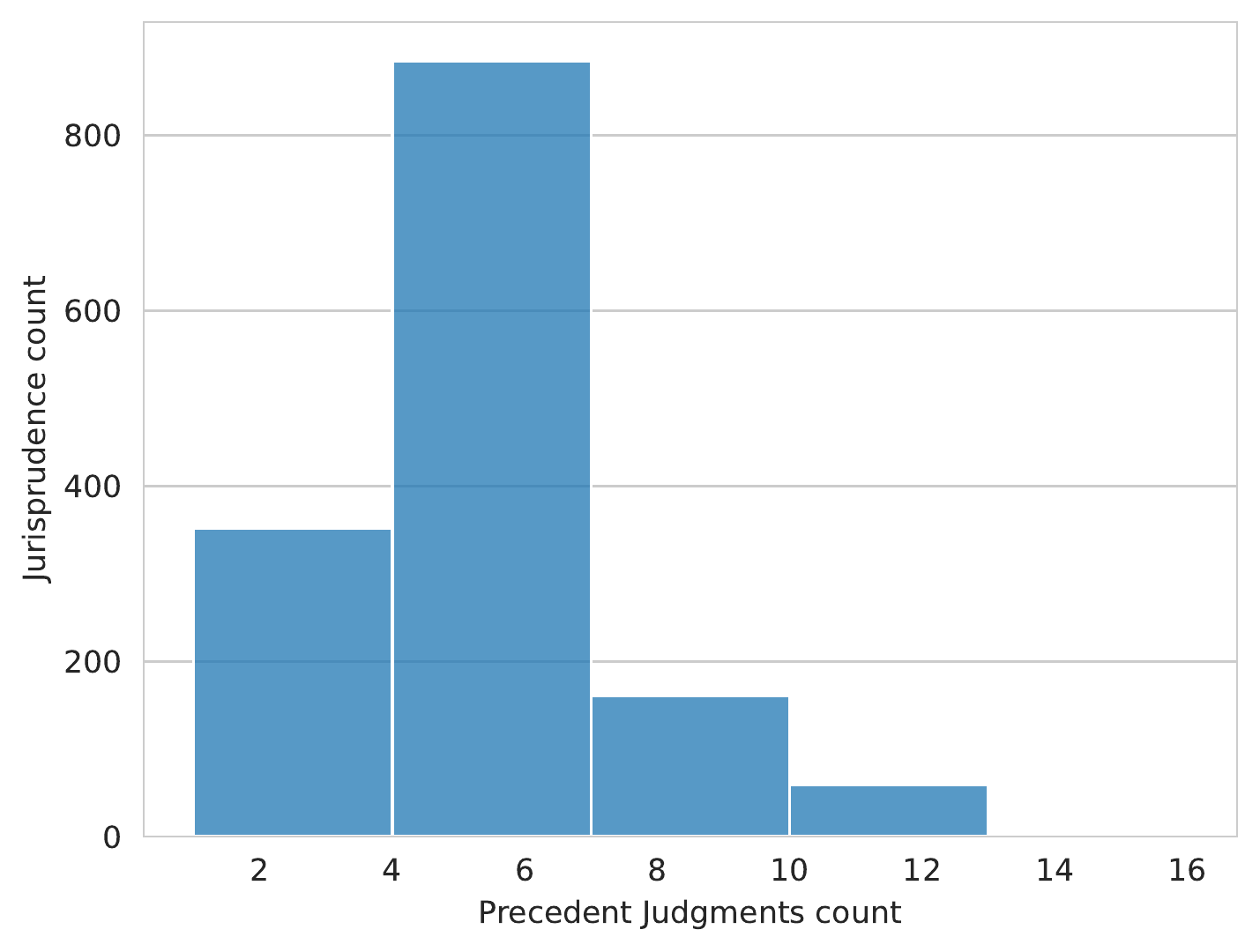}
\caption{Histogram of Precedents x \emph{STJ decisions} Jurisprudences.}
\label{fig:prec_juris_stj}

\endminipage
\end{figure}

The graphs that explore the composition of the aforementioned datasets are presented next. Figure \ref{fig:prec_juris_tcu} presents a histogram of the \emph{TCU votes} dataset, indicating that in this dataset, jurisprudence has, on average, between seven and eight previous votes. On the other hand, the histogram presented in Figure \ref{fig:prec_juris_stj} related to \emph{STJ decisions} shows that most case law has between five and six precedent decisions.

\begin{figure*}[!h]
\centering
\includegraphics[width=\linewidth]{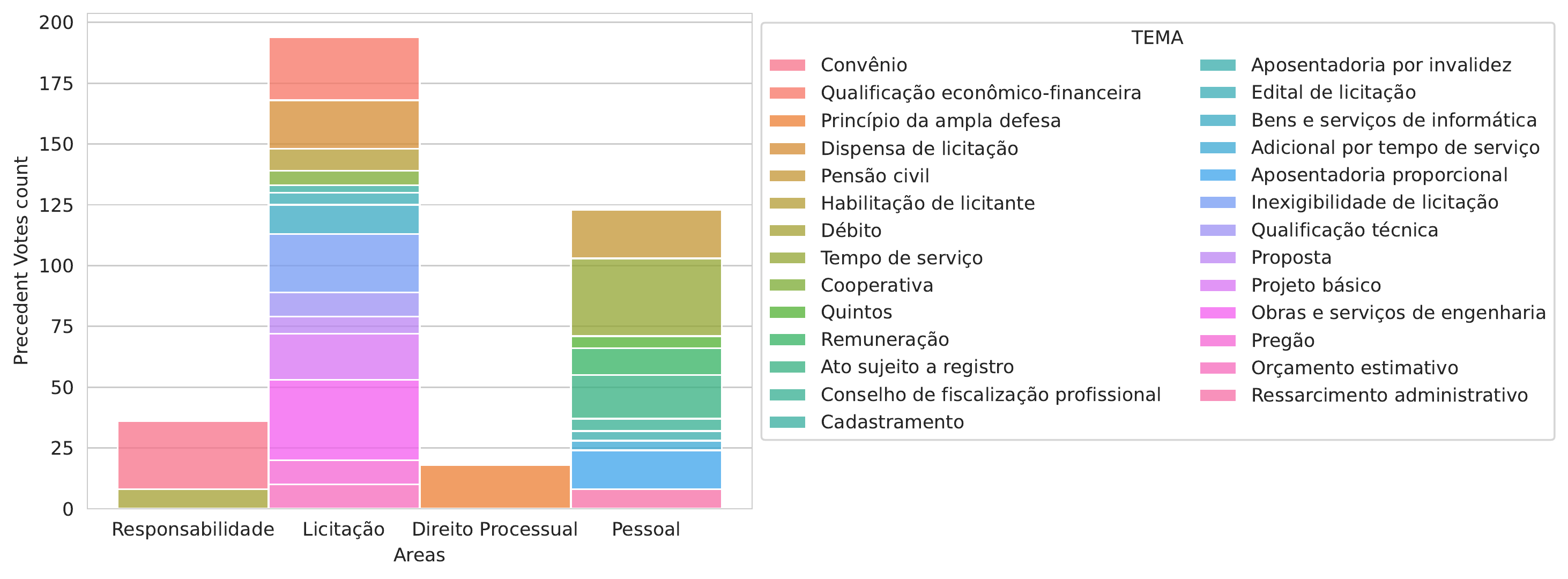}
\caption{Histogram of VOTE x AREA x THEME of \emph{TCU votes}}
\label{fig:prec_detalhe_tcu}
\end{figure*}

\begin{figure*}[htbp]
\centering
\includegraphics[width=\linewidth]{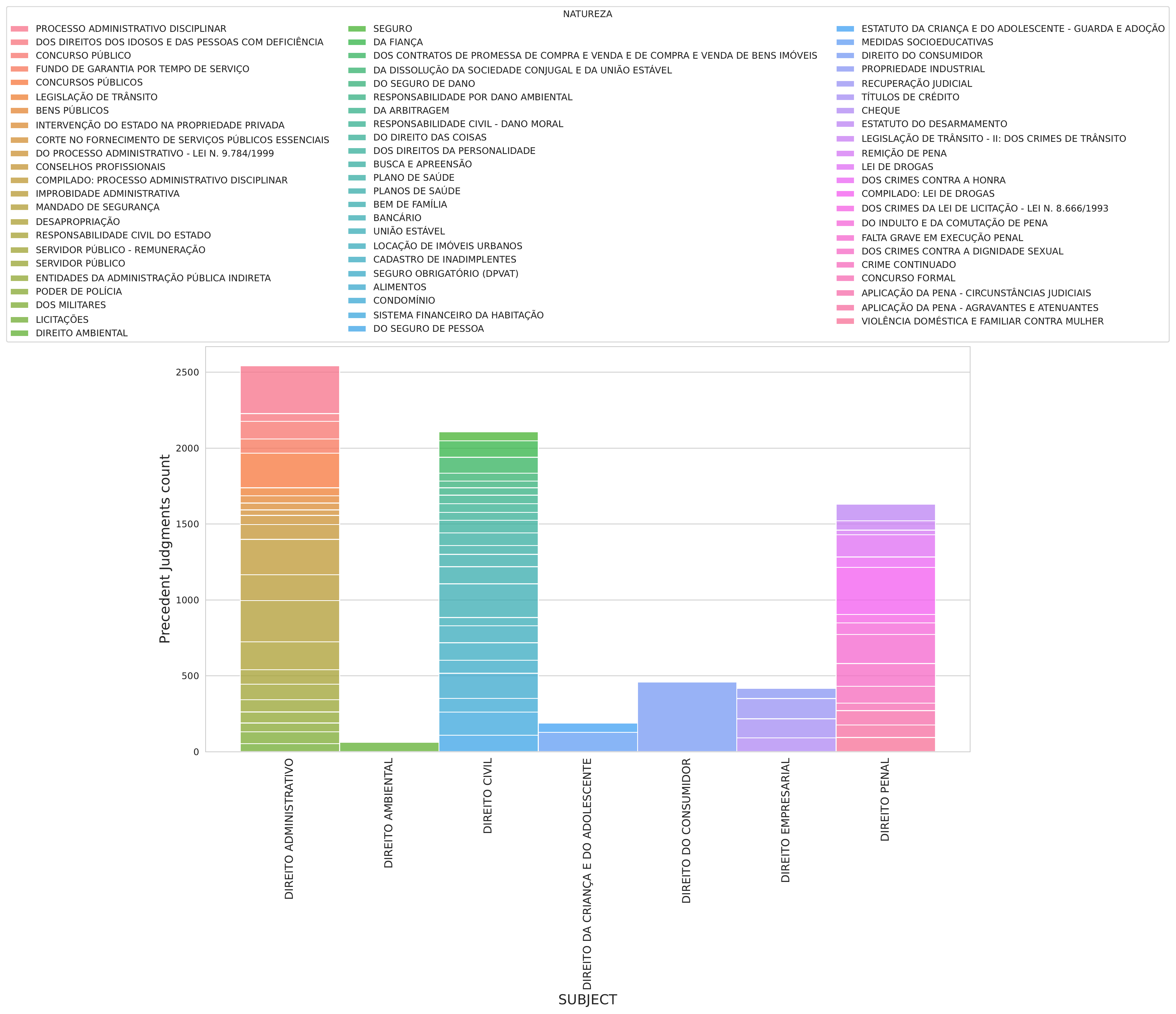}
\caption{Histograms of DECISION x MATTER x NATURE of \emph{STJ decisions}}
\label{fig:prec_detalhe_stj}
\end{figure*}

Figure \ref{fig:prec_detalhe_tcu} shows that the precedents of the  \emph{TCU votes} dataset are primarily from the {\itshape LICITAÇÃO} area, followed by the Personal area. Furthermore, the Bidding and Personal areas have the greatest dispersion of precedents across different Themes. With regard to the \emph{STJ decisions}, Figure \ref{fig:prec_detalhe_stj} shows that the precedents are mainly of the Subjects: Administrative Law, Civil Law, and Criminal Law. The dispersion by Nature of precedents in these three Matters is also more significant than in the others.

\begin{figure*}[htbp]
\minipage{0.49\linewidth}

\includegraphics[width=\linewidth]{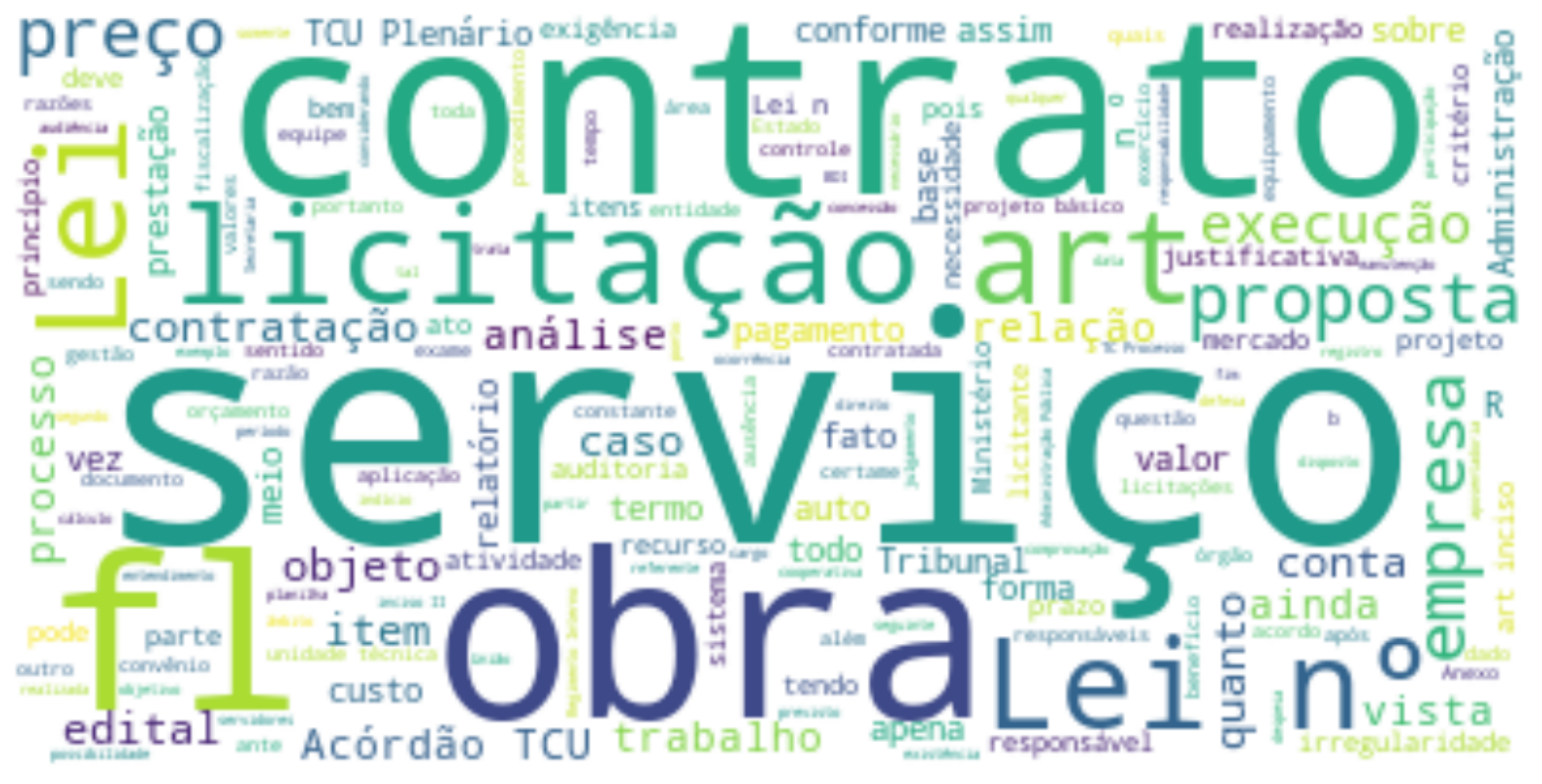}
\caption{Wordcloud \emph{TCU votes} precedents.}
\label{fig:nuvem_precedentes_tcu}

\endminipage\hfill
\minipage{0.49\textwidth}

\includegraphics[width=\linewidth]{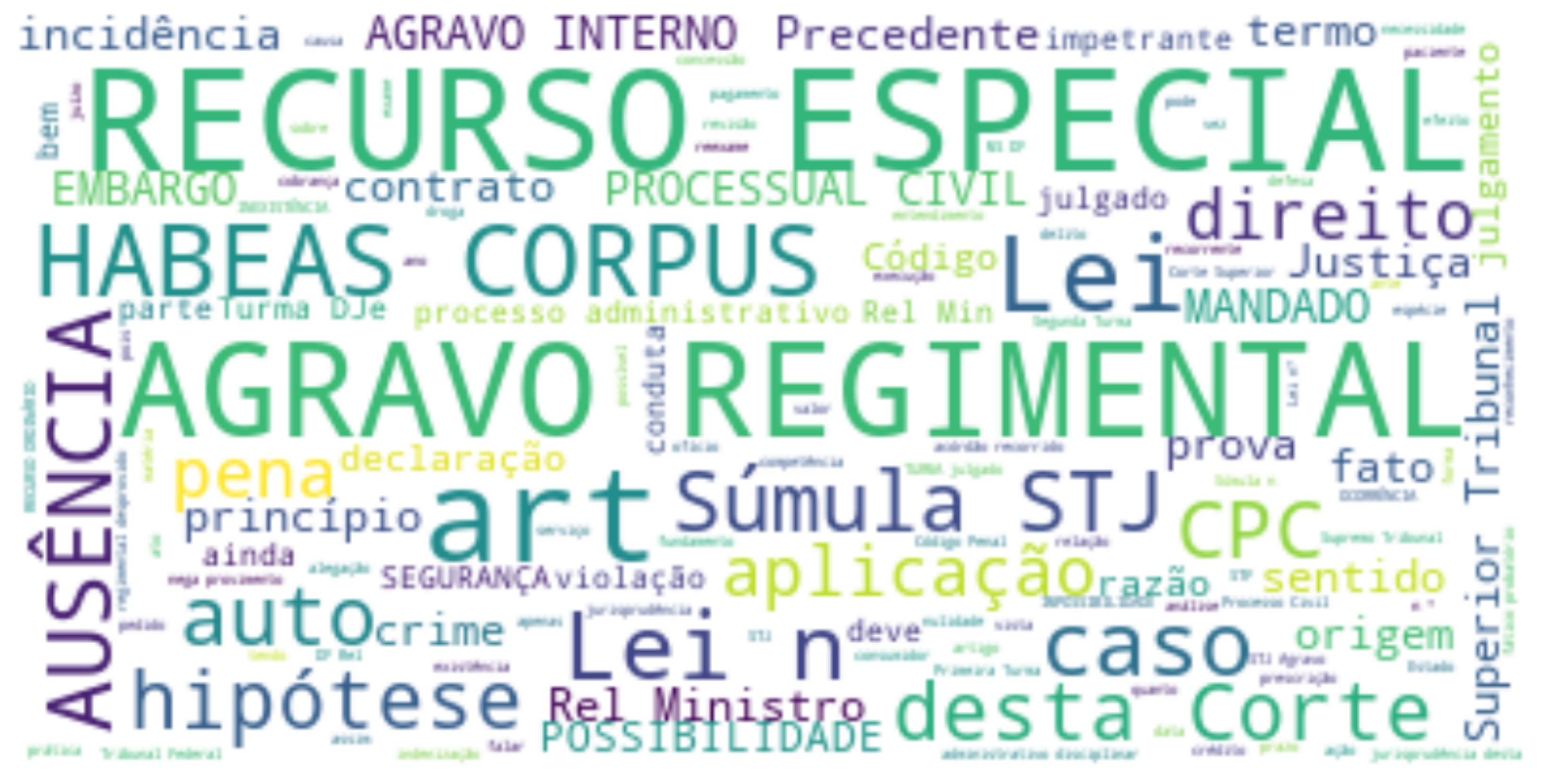}
\caption{Wordclound \emph{STJ decisions} precedents}
\label{fig:nuvem_precedentes_stj}

\endminipage
\end{figure*}

The word cloud of the \emph{TCU votes} in Figure \ref{fig:nuvem_precedentes_tcu} highlights words like \texttt{WORK, SERVICE, CONTRACT, and BIDDING} as the most frequent in the precedents of the dataset. Meanwhile, Figure \ref{fig:nuvem_precedentes_stj} highlights terms such as \texttt{SPECIAL APPEAL, REGULATORY APPEAL, HABEAS CORPUS, CIVIL PROCEDURE, and INTERNAL APPEAL} as the most frequent in the dataset \emph{STJ decisions}.

\begin{figure*}[htbp]
\centering
\includegraphics[width=\linewidth]{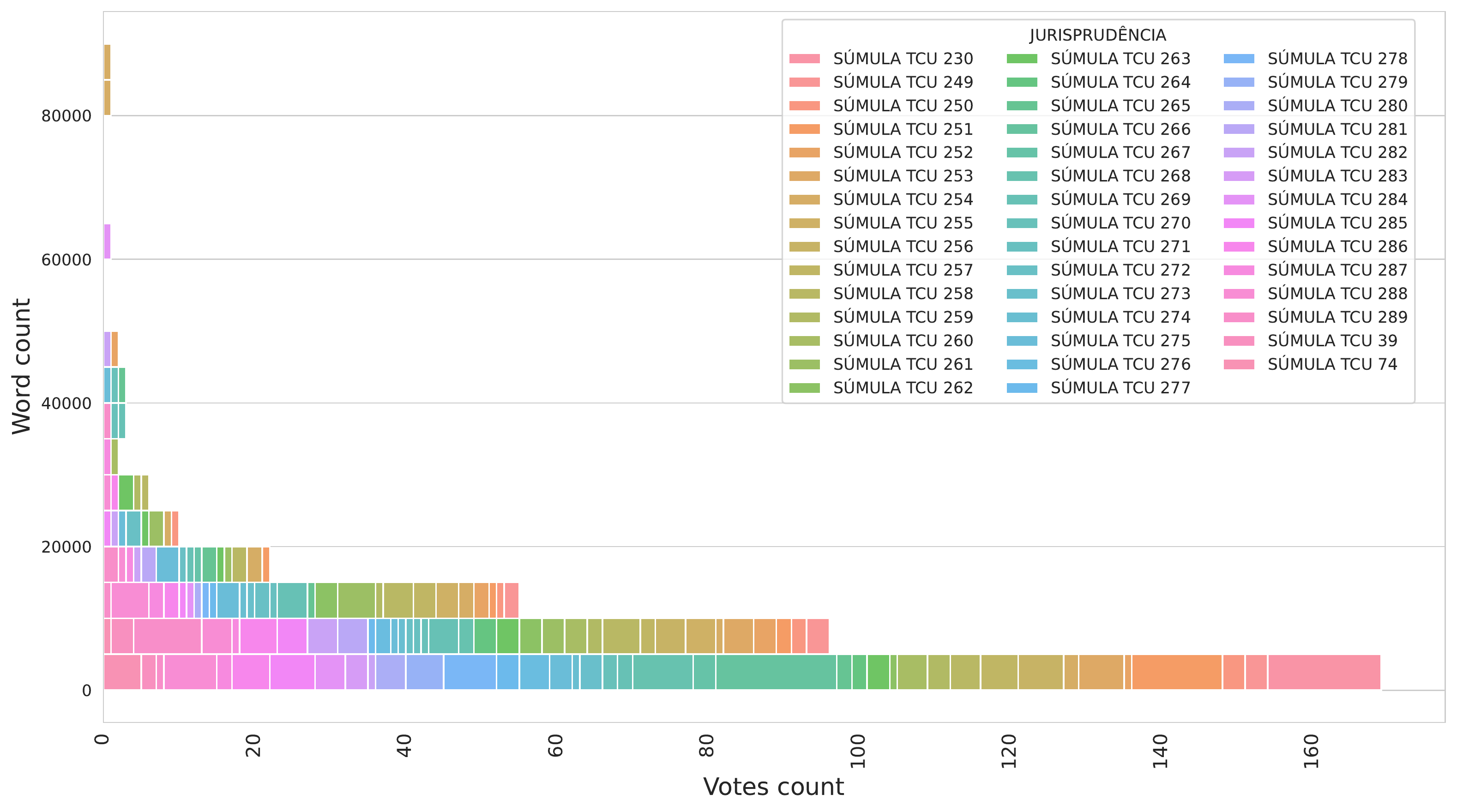}
\caption{Histogram of words x \emph{TCU votes} precedents}
\label{fig:palavras_tcu}
\end{figure*}

\begin{figure*}[htbp]

\includegraphics[width=\linewidth]{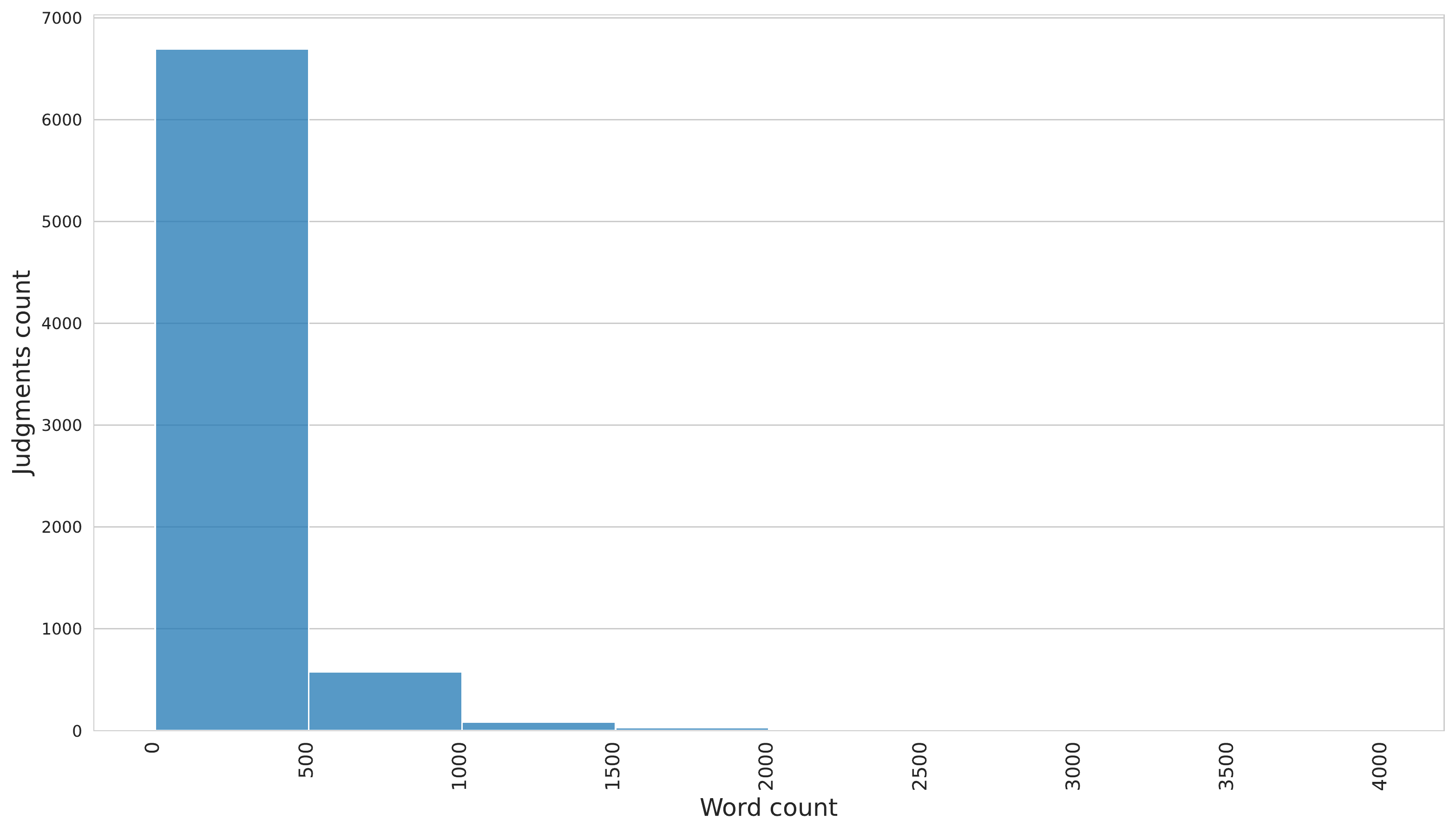}
\caption{Histogram of words x \emph{STJ decisions} precedents.}
\label{fig:palavras_stj}
\end{figure*}

The histogram in Figure \ref{fig:palavras_tcu} indicates that most precedents in the \emph{TCU votes} dataset have up to 20,000 words. In this case, the words are defined from the break by spaces in the texts of the precedents. When considering the \emph{STJ decisions}, Figure \ref{fig:palavras_stj} shows that most of the precedents in this dataset have up to 500 words.

\section{An Heuristics for annotating Legal Semantic Textual Similarity Datasets} \label{sec:corpussst}

Given the importance of recovering similar processes in the legal context and the absence of datasets that help in the process of training models for the Semantic Textual Similarity task, the main contribution of this article are the datasets \emph{TCU votes for Semantic Textual Similarity} and \emph{STJ decisions for Semantic Textual Similarity}. They were constructed from the datasets presented in Section \ref{sec:corpus} and synthesized for the  Semantic Textual Similarity (STS) task~\cite{fonseca2016assin} . Commonly, an STS  dataset consists of a pair of texts and a score associated with their semantic similarity. The higher the score, the greater the semantic similarity between the texts. 

Datasets for the STS task are expensive to prepare, as they require the mobilization of human users to annotate the data, and, in many cases, they need to be experts in the data domain. This way, this article contributes with an automatic process based on a heuristic derived from the metadata of the texts to alleviate the need for human annotators in STS tasks. Given the particular nature of both datasets, we propose two separate processes to annotate the datasets from STJ and TCU. Accordingly, to annotate \emph{STJ decisions for Semantic Textual Similarity} dataset, the procedure performs the following steps, given the hierarchical order existing between the documents (Table \ref{tab:data_info}):

\begin{enumerate}
   \item Generate pairs between decisions of the \emph{same Jurisprudence} and assign to each pair a score with a base value of $4.5$, plus a noise that obeys a normal distribution among all generated pairs.
   \item Generate pairs between decisions of the \emph{same Nature} and assign a score to each pair with a base value of $3$, plus a noise that obeys a normal distribution among all generated pairs.
   \item Generate pairs between decisions of \emph{different Matters} and assign to each pair a score with a base value of $0.5$, plus a noise that obeys a normal distribution among all generated pairs.
   \item Generate the final set with a \emph{balanced combination} of previously generated subsets.
\end{enumerate}

The heuristic to annotate the \emph{STJ decisions for Semantic Textual Similarity} dataset assumes that decisions that served as precedents for the same Jurisprudence have a great intrinsic similarity. On the other hand, decisions that are of the same Nature, but are not precedents of the same Jurisprudence, keep a less intense similarity relationship. Finally, judgments dealing with different Matters are quite different. This last set does not contain documents from different jurisprudence because although they are not precedents from the same jurisprudence, they may have the same Nature and thus retain some degree of similarity. The choice of three base values, $(4.5 , 3, 0.5)$, was to simulate pairs of documents with high similarity, neutral, or dissimilarity. The inclusion of noise following a normal distribution aimed to simulate the uncertainty and difference between annotations when a manual annotator performs the process.

The procedure to annotate the \emph{TCU votes for Semantic Textual Similarity} is similar to the previous one, except for the differences in the metadata used. In this case, the following steps are followed:

\begin{enumerate}
   \item Generate pairs between votes of the \emph{same Jurisprudence} and assign to each pair a score with a base value of $4.5$, plus a noise that obeys a normal distribution among all generated pairs.
   \item Generate pairs between votes from the \emph{same Area and Theme} and assign to each pair a score with a base value of $3$, plus a noise that obeys a normal distribution among all generated pairs.
   \item Generate pairs between votes from \emph{different Areas} and assign each pair a score with a base value of $0.5$, plus a noise that obeys a normal distribution among all generated pairs.
   \item Generate the final set with a balanced combination of previously generated subsets.
\end{enumerate}

A striking difference when labeling the \emph{TCU votes for Semantic Textual Similarity} dataset compared to the former procedure is the second subset, which regards votes from the same Area and Theme. The data scraped from TCU contain Themes with the same terminology but belonging to different Areas. 

The TCU dataset has $4,843$ tuples, while the STJ dataset has $51,437$ tuples. After the automatic process of generating pairs and associated scores, and balancing between the subsets generated in each step, we further divide them into \texttt{TRAINING}, \texttt{TEST} and \texttt {VALIDATION} sets, holding the proportion of pairs per similarity interval. Thus, the dataset for STS with TCU votes was divided into $3,389$ for training, $438$ for validation, and $1,016$ for testing. The dataset for STS with STJ decisions was divided into $36,010$ for training, $4,613$ for validation, and $10,814$ for testing.

\section{Building datasets for the Legal STS task with expert-annotation labels}\label{sec:ground_truth}

We collected labels from expert annotators and compared them to the heuristic-induced labels to evaluate the effectiveness of the heuristic proposed in this article.
The annotation was conducted with the help of Google Forms \footnote{https://www.google.com/forms/about/}. Document pairs were presented to legal domain experts, and six questions were asked about those:

\begin{enumerate}
   \item How semantically similar are the two documents?
   \item What is your level of confidence in the assigned similarity?
   \item Which part of the first document was most relevant for the attributed similarity?
   \item Was the most relevant part of the first document in the header (initial part in capital letters) or in the body?
   \item Which part of the second document was most relevant for attributed similarity?
   \item Was the most relevant part of the second document in the header (initial part in capital letters) or in the body?
\end{enumerate}

The first question is the most important to make us able to evaluate the heuristic labeling. The annotators must select one between five options elaborated in a Likert scale \cite{joshi2015likert} with values in the range $[0,4]$:

\begin{itemize}
   \item 0 - Not related
   \item 1 - A little related
   \item 2 - Somewhat related but not similar
   \item 3 - A little similar
   \item 4 - Very similar
\end{itemize}

The annotators used a guide to help them choose the value, the same one used in \cite{cer-etal-2017-semeval}, where scenarios are shown where each label is likelier to be used. The second question is intended to measure the uncertainty that even a expert annotator can have and can help to evaluate the heuristic labeling. The results of the third and fifth questions can enhance supervised Machine Learning methods by highlighting the relevant parts of the documents evaluated in the STS task. The fourth and sixth questions want to evidence if document structure that can be easily extracted can be used to enhance the heuristic labeling process. 

As mentioned before, access to expert annotators is challenging, and, therefore, the following experiment and results were performed using only the STJ decisions. The option to use STJ decisions instead of TCU votes is because the TCU body has a broader occupation area of knowledge, which implicates law subfields experts to evaluate the document pairs. The domain experts were 27 students of the Master's in Law course and were invited to answer the questions in the classroom. We initially selected 140 document pairs from the STJ dataset test set to be annotated. Therefore, fourteen \textit{Google Form} were created, each with ten document pairs. Initially, we were supposed to use thirteen forms, where two experts annotated each, and the remaining would be annotated by only one. That would give us $270$ labeled document pairs for the STS task. However, at the end of the form assignment, we collected $240$ labeled document pairs, where two domain experts labeled $100$ unique document pairs and only one domain expert labeled $40$ document pairs.

Then we investigate five research questions:

\begin{enumerate}
\item How related are the domain experts' labels for the same pair of documents?
\item Where are the highlight parts used to annotate the documents, body, or header?
\item What is the distribution of the labels in the dataset annotated by the domain experts?
\item How related are the domain expert and heuristic labels?
\item What is the mean and standard deviation of the domain experts' confidence in the assigned labels?
\end{enumerate}

\begin{figure}[!htb]
\centering
\minipage{0.49\textwidth}

\includegraphics[width=\textwidth]{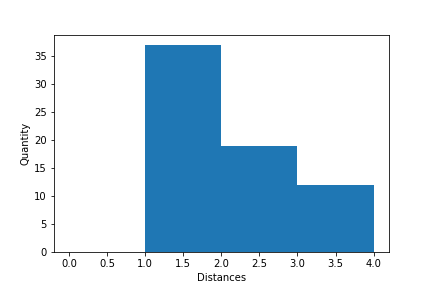}
\caption{Histogram of the distance between labels, between the data labeled in divergence}
\label{fig:histogram_distance_divergent}
\endminipage\hfill
\end{figure}

To answer the first research question, we considered the $100$ unique document pairs labeled by two domain experts and found that only $32$ of the unique document pairs were equally labeled by two experts. Then, we investigate the $68$ document pairs that were labeled divergently by two domain experts. We calculated the distance between the divergent labels for each document pair and the mean, variance, and standard deviation based on these distances. We found that the mean distance between divergent labels was $1.63$, the variance $0.58$, and the standard deviation $0.76$. We also calculated the Pearson and Spearman correlation beyond Krippendorff's alpha \citep{krippendorff2004reliability} although the last one would be more interesting if we have more than two labels by document pair. Pearson correlation was $0.63$, and the Spearman correlation was $0.60$, which suggests a positive correlation between the divergent labels, and makes sense since all labels are positive numbers and there is a small range. Krippendorff's alpha was $-0.12$, which makes sense since negative values imply less concordancy than would be in a chance scenario, and this is the case since all labels are divergent.
Figure \ref{fig:histogram_distance_divergent} shows the distribution of distances between divergent labeled document pairs.

We evaluate the divergency distances homogeneously. However, in this scenario of semantic similarity labeling using a Likert scale, some divergence can be seen as a \textit{partial divergence} and others as a \textit{total divergence}. A document pair where a domain expert gives a \textit{1 - A little related} label and another domain expert gives a \textit{0 - Not related} can be seen as a \textit{partial divergence} since neither sees the document pair similar. However, in the case where a domain expert gives a \textit{0 - Not related} label and another domain expert gives a \textit{3 - A little similar} label, it can be seen as a \textit{total divergence}. We also investigate this in detail and categorize the cases where label divergence for document pairs is in $[0, 1, 2]$ or in $[2, 3, 4]$ as a case of \textit{partial divergence}. In cases where one domain expert assigns a label in $[0, 1]$ and another in $[3,4]$, there is a \textit{total divergence}.

\begin{table}[]
\centering
\begin{tabular}{|l|l|l|}
\hline
                                         & \textbf{Total} & \textbf{\begin{tabular}[c]{@{}l@{}}Mean \\ distance\end{tabular}} \\ \hline
Partial divergences about non-similarity & 28             & 1.11                                                              \\ \hline
Partial divergences on similarity        & 18             & 1.33                                                              \\ \hline
Total divergences                        & 22             & 2.54                                                              \\ \hline
\end{tabular}
\caption{Divergences in a stratified mode}
\label{tab:divergence_analysis}
\end{table}

Table \ref{tab:divergence_analysis} shows that between the $68$ unique document pairs divergently labeled by two experts, $22$ is a total divergence. This total divergence represents more than $20\%$ of divergence between the $100$ unique document pair labeled by real annotators, which helps to state the difficulty of the task of Semantic Textual Similarity in a Legal domain. 

An example of document pair with total divergent labels where a domain expert assigns a similarity score of $0$, and another assigns $3$ is:

\blockquote{
\textbf{Document 1}

ADMINISTRATIVO  E  PROCESSUAL  CIVIL.  AGRAVO  INTERNO  NO AGRAVO EM
RECURSO  ESPECIAL.  ACÓRDÃO  A  QUO  QUE DIRIMIU TODA A CONTROVÉRSIA
POSTA  NOS  AUTOS.  FUNDAMENTAÇÃO  SUFICIENTE. NEGATIVA DE PRESTAÇÃO
JURISDICIONAL.  NÃO  OCORRÊNCIA. AÇÃO CONSUMERISTA. INVERSÃO DO ÔNUS
DA  PROVA  EM  FAVOR  DO  PARQUET.  POSSIBILIDADE.  1. Nos termos da
orientação   jurisprudencial   deste   Superior  Tribunal,  tendo  a
instância de origem se pronunciado de forma clara e precisa sobre as
questões   aventadas   no   feito,   assentando-se   em  fundamentos
suficientes  para  embasar  a  decisão,  não  há falar em omissão no
acórdão  regional, uma vez que a fundamentação sucinta não significa
ausência de fundamentação.
2. Na hipótese dos autos, não ocorreu a alegada ofensa ao art. 1.022
do  CPC/2015,  na  medida  em  que  o  Tribunal  de  origem dirimiu,
fundamentadamente,  as questões que lhe foram submetidas, apreciando
integralmente  a  controvérsia  posta  nos  autos,  não  se podendo,
ademais, confundir julgamento desfavorável ao interesse da parte com
negativa ou ausência de prestação jurisdicional.
3.  Acerca  da inversão do ônus da prova, a Corte local alinhou-se à
jurisprudência  deste  Sodalício  sobre  o  tema,  cujo entendimento
assevera  que,  "na  ação  consumerista  deflagrada  pelo Ministério
Público,  não  se  indaga  de  hipossuficiência do demandante para a
inversão  do  ônus  da  prova,  pois  a  presença  do  Parquet  como
substituto  processual  da coletividade assim o justifica" (AgInt no
AREsp 222.660/MS, Rel. Ministro Gurgel de Faria, Primeira Turma, DJe
19/12/2017).
4. Agravo interno a que se nega provimento.
}

\blockquote{
\textbf{Document 2}

AGRAVO REGIMENTAL. PROCESSUAL CIVIL. NÃO HÁ QUE SE FALAR EM VIOLAÇÃO
DO ARTIGO 535 DO CÓDIGO DE PROCESSO CIVIL QUANDO O ACÓRDÃO DIRIME,
FUNDAMENTADAMENTE, AS QUESTÕES PERTINENTES AO LITÍGIO. NOS TERMOS DA
SÚMULA 283 DO SUPREMO TRIBUNAL FEDERAL, QUANDO A DECISÃO RECORRIDA
TEM POR BASE MAIS DE UM FUNDAMENTO, DEVE O RECURSO ABRANGER TODOS
ELES. A SÚMULA 60 DESTA CORTE ORIENTA SER NULA A OBRIGAÇÃO CAMBIAL
ASSUMIDA POR PROCURADOR DO MUTUÁRIO VINCULADO AO MUTUANTE, NO
EXCLUSIVO INTERESSE DESTE. ORIENTA A SÚMULA 83 DESTE TRIBUNAL, QUE
NÃO SE CONHECE DE RECURSO FUNDADO EM DIVERGÊNCIA QUANDO ORIENTAÇÃO
DESTA CORTE SE FIRMOU NO MESMO SENTIDO DA DECISÃO RECORRIDA. .
APLICAÇÃO DA MULTA PREVISTA NO ARTIGO 557, § 2º, DO CÓDIGO DE
PROCESSO CIVIL. AGRAVO IMPROVIDO.
}

To understand what can influence the divergent annotation behavior in scenarios like this, the document pair example and the two previous labels were presented to a third domain expert to examine. According to the domain expert, the label \textit{3 - A little similar} only holds up because both documents contain a procedural issue specific to the appeal before the STJ regarding the grounds for the appeal. However, the issues discussed on the merits of the appeal are entirely different, which justifies the label \textit{0 - Not related}. Those who labeled as $3$ took into account this initial debate of a procedural nature. Nevertheless, all the appeals must examine this, so it is not very useful to distinguish if everyone has to check these issues. Label $0$ is more interesting because that considers the appeals' fundamental issues.

In response to research question 2, we found that text portion positions are irrelevant to determine if they have more chance to be a highlight of a document because the text highlighted positions are homogenously distributed between the header and body of documents in the ground truth dataset. Also, $83\%$ of the time, the highlighted text position of document 1 differs from that of document 2.

\begin{figure}[!htb]
\centering
\minipage{0.49\textwidth}

\includegraphics[width=\textwidth]{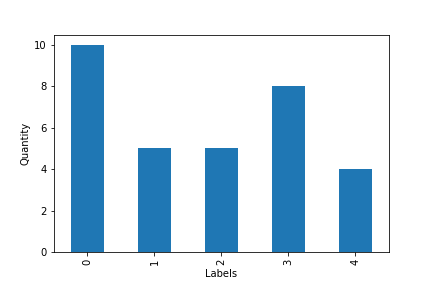}
\caption{Domain expert labels distribution}
\label{fig:expert_label_distribution}
\endminipage\hfill
\end{figure}

With the help of Figure \ref{fig:expert_label_distribution}, we can respond to the third research question. Figure \ref{fig:expert_label_distribution} shows that the labels assigned by the domain experts are majority 'Not related' between the documents, followed by 'A little similar'. Since the selected document pairs to be annotated came from the heuristic dataset, the heuristic can minimally distinguish some relation similarity and dissimilarity of texts.

The fourth research question is critical since its answer implies a direct measure of the heuristic's utility for labeling a dataset for STS with the STJ data. Only the $32$ pairs of documents equally labeled by two domain experts were used to answer the fourth research question. We calculated Pearson and Spearman correlations and Krippendorff's alpha using the labels from the heuristic and ground truth datasets. The result for Pearson correlation was $0.45$, for Spearman correlation was $0.43$, and for Krippendorff's alpha was $0.40$. The results of correlation metrics can be interpreted as a moderate positive correlation between the label generated by the heuristic and the ground truth \citep{evans1996straightforward}\citep{altman1990practical}.

Finally, to answer the fifth research question, we processed the responses about the self-confidence of the domain expert in the given annotation. The answers about the confidence were retrieved using a Likert scale in $[0, 4]$, and the mean confidence was $3.28$, with a variance of $0.93$ and $0.96$ standard deviation.

The datasets generated and/or analysed during the current study are available in \url{https://osf.io/mct8s/}. The source code used is available in \url{https://github.com/danieljunior/jidm}.

\section{Conclusions}\label{sec:conclusion}

This article contributes with two datasets of the legal domain, a heuristic process to generate labeled datasets for the Semantic Textual Similarity task, two datasets heuristic labeled to be used in the Semantic Textual Similarity task, and a small ground truth dataset from a subset of one of the heuristic labeled dataset. The first two datasets from the legal domain were built from data collected from the Federal Court of Auditors (TCU) and the Superior Court of Justice (STJ) websites. The data collection generates the \emph{TCU votes} and \emph{STJ decisions}  datasets related to case law precedents. Beyond the precedent textual content, it also has metadata related to the documents' categorizations in the context of the respective bodies.

The main contribution of this article is the proposal of a heuristic for automatically annotating the datasets for the Semantic Textual Similarity task with legal domain data. With the help of the proposed heuristic process, this article also makes available two heuristic-labeled datasets \emph{TCU Votes for Semantic Textual Similarity} and \emph{STJ decisions for Semantic Textual Similarity}, constructed from the collected precedents and the application of the heuristic. In addition to the legal domain dataset build and availability and the creation of the heuristic, this article contributes with an exploratory analysis of these sets.

The effectiveness of the proposed heuristic annotation was evaluated with the help of a ground truth dataset generated through a data annotation process with legal domain experts designed as a question-answer experiment. This experiment highlights that the specific domain annotation of semantic textual similarity can generate relevant divergences labeling between domain experts, making the difficulty of automating such a process more explicit. Finally, comparing the heuristic labels and ground truth labels was found that the heuristic process can be used with moderate confidence on the generated labels. Future work includes using datasets proposed here to evaluate unsupervised methods for legal document retrieval. Datasets proposed also allow the opportunity to adapt and generate Machine Learning models for the Portuguese Legal Semantic Textual Similarity task.

\bibliographystyle{unsrtnat}
\bibliography{references}  






\end{document}